\pdfoutput=1

\documentclass[11pt]{article}

\usepackage{acl}

\usepackage{times}
\usepackage{latexsym}

\usepackage[T1]{fontenc}

\usepackage[utf8]{inputenc}

\usepackage{microtype}

\usepackage{amsmath}
\usepackage{amssymb}
\usepackage{multirow}

\usepackage{pifont}
\newcommand{\xmark}{\ding{55}}%

\usepackage{graphicx}
\usepackage{svg}

\title{
Fine-Tuning Pre-trained Transformers into Decaying Fast Weights
}

\author{Huanru Henry Mao \\
  Jenni \\
  \texttt{henry@jenni.ai}
 }

\begin{document}
\maketitle
\begin{abstract}
Autoregressive Transformers are strong language models but incur $O(T)$ complexity during per-token generation due to the self-attention mechanism.
Recent work proposes kernel-based methods to approximate causal self-attention by replacing it with recurrent formulations with various update rules and feature maps to achieve $O(1)$ time and memory complexity.
We explore these approaches and find that they are unnecessarily complex, and propose a simple alternative - \textbf{decaying fast weights} - that runs fast on GPU, outperforms prior methods, and retains 99\% of attention's performance for GPT-2.
We also show competitive performance on WikiText-103 against more complex attention substitutes.
\end{abstract}

\section{Introduction}
Autoregressive Transformers \cite{vaswani2017attention} have demonstrated strong performance on text generation \cite{brown2020language}.
The success of self-attention in Transformers over recurrent models \cite{hochreiter1997long} can be attributed to its parallelizability \cite{hooker2021hardware} and its effective gradient propagation over many time steps \cite{ke2018sparse}.
However, self-attention has a high computation and memory cost. During inference sampling, it consumes $O(T)$ time and memory and grows linearly per token generated.

These drawbacks motivated recent work to \textit{convert} or \textit{fine-tune} attention into recurrent formulations with $O(1)$ memory and time complexity for auto-regressive generation.
\textbf{Kernel-based methods} for self-attention \cite{tay2020long} learn approximations of the exponential similarity function using $m$-dimensional feature maps to reformulate attention as a recurrent computation.
They replace attention with ``unlimited capacity'' with fixed-capacity fast weights \cite{schmidhuber1992learning,peng2021abc}, where the memory-accuracy trade-off \cite{kerg2020untangling} is controlled by $m$.
Several works explored different feature maps and recurrent formulations (i.e.,~update rules).
\citet{katharopoulos2020transformers} propose feature maps to maintain positive outputs, while \citet{choromanski2020rethinking,peng2021random} carefully ensure their \textit{random} feature maps are unbiased estimates of the softmax attention kernel.
\citet{schlag2021linear,peng2021random} propose more sophisticated update rules to \textit{forget} information in the recurrent state to improve performance.
Recently, \citet{kasai2021finetuning} showed that pre-trained Transformers can be fine-tuned into a recurrent formulation using learned ReLU feature maps with minor degradations.
While promising, it is unclear which update rules or feature maps are critical for successful fine-tuning.

\begin{figure}[t]
    \centering
    \includegraphics[width=\columnwidth]{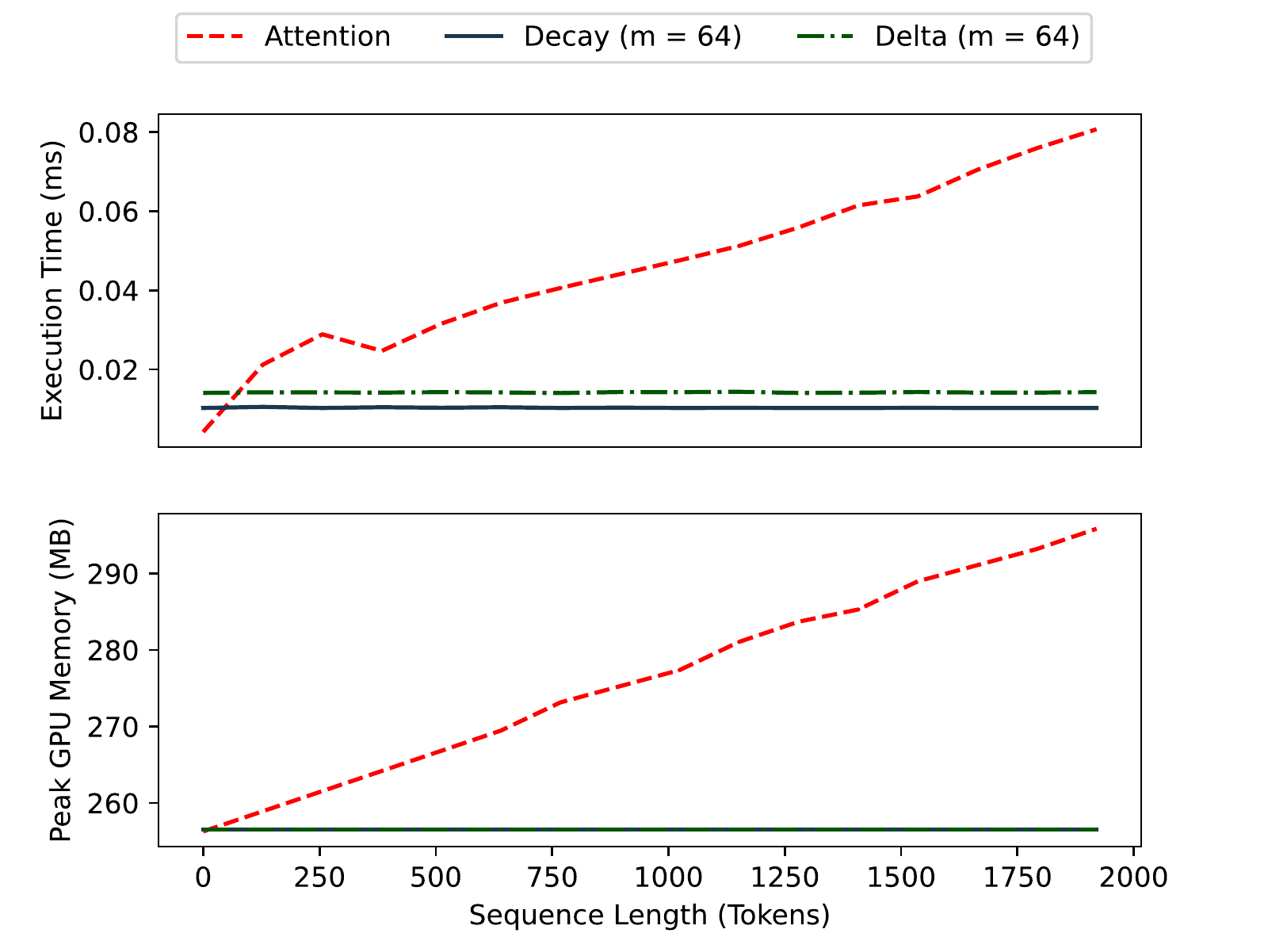}
    \caption{
        Plot of memory usage and execution time of \textbf{our decay rule}, \textbf{delta rule} and \textbf{attention} when generating the next token at various sequence lengths on Quadro RTX 4000.
        Decay and delta rule use approximately the same peak memory (overlapped in plot).
    }
\label{fig:profiling}
\end{figure}

In this work, we investigate various update rule configurations to fine-tune pre-trained Transformers into RNNs for fast inference.
We find that \textbf{prior proposals contain unnecessary operations}, leading us to propose a simple element-wise \textbf{decay update rule} with \textbf{no feature map}.
We fine-tune GPT-2 \cite{radford2019language} into our recurrent formulation to demonstrate that our rule outperforms prior methods and \textbf{recovers 99\% of self-attention's performance}.
We also show competitive performance on WikiText-103 \cite{merity2016pointer} compared to more complex attention alternatives.
Our results support the idea \cite{merity2019single,zhai2021attention} that it is unnecessary for attention alternatives to maintain a close analogy to self-attention, and it is more important to focus on designing an expressive update rule.

\section{Background and Related Work}

\subsection{Kernel-based Self-Attention}
\label{sec:attention}
Kernel-based approximations \cite{katharopoulos2020transformers,choromanski2020rethinking,kasai2021finetuning} to self-attention reorders computation such that a typical $O(Td)$ (per token) memory and time complexity attention becomes $O(dm)$ for $T$ time steps, dimension $d$ and feature size $m$.
Given input to the attention layer $x_t \in \mathbb{R}^{d \times 1}$ and learned weight matrices $W_*$, the causal self-attention \cite{vaswani2017attention} for query $q_t = W_q x_t \in \mathbb{R}^{d \times 1}$, key $k_t  = W_k x_t \in \mathbb{R}^{d \times 1}$ and value $v_t  = W_v x_t \in \mathbb{R}^{d \times 1}$ is defined as:
\begin{equation}
\begin{aligned}
\label{eq:attention}
y_t = \sum_{j}^{t} \frac{ \text{sim} (k_j,q_t) } { \sum_{i}^{t} \text{sim} (k_i, q_t) } v_j \\
\text{sim}(x,y) = \exp (x^\intercal y / \sqrt{d})
\end{aligned}
\end{equation}
Kernel-based methods propose an approximation to the exponential similarity function $\widetilde{sim}(x,y)= \phi(x)^\intercal \phi(y)$ via a $m$-dimensional kernel feature map $\phi: \mathbb{R}^d \xrightarrow{} \mathbb{R}^m$. This approximation enables us to rewrite Eq. \ref{eq:attention} as
\begin{equation}
y_t =\frac{ \sum_{j}^{t} v_j \phi(k_j)^\intercal \phi(q_t) } { \sum_{i}^{t} \phi(k_i)^\intercal \phi(q_t) }
\end{equation}
due to the associative property of matrix multiplication.
This lends itself to a recurrent formulation with state $S_t \in \mathbb{R}^{d \times m}$ and normalizer $z_t \in \mathbb{R}^{1 \times m}$ that can be computed at every time step: 
\begin{equation}
\label{eq:attention-rec-add}
S_{t} = S_{t-1} + v_t \phi(k_t)^\intercal, z_{t} = z_{t-1} + \phi(k_t)^\intercal
\end{equation}
The state recurrence resembles fast weight additive outer products \cite{schmidhuber1992learning}.
Finally, the output is computed by normalizing against $z_t \phi(q_t)$, which we refer to as \textbf{attention normalization}:
\begin{equation}
\label{eq:attn-norm}
y_t = \frac{S_t \phi(q_t)} {z_t \phi(q_t)}
\end{equation}
\subsection{Update Rules}
Because Eq. \ref{eq:attention-rec-add} is an \textbf{additive update rule}, it is unable to forget past memories.
This can overwhelm the fixed state capacity and lead to poor performance.
\citet{peng2021random} proposes a \textbf{gated rule} similar to gated RNNs \cite{chung2014empirical} to decay old information and induce a recency bias:
\begin{equation}
\label{eq:attention-rec-peng}
\begin{aligned}
S_{t} = g_t S_{t-1} + (1 - g_t) v_t \phi(k_t)^\intercal \\
z_{t} = g_t z_{t-1} + (1 - g_t) \phi(k_t)^\intercal
\end{aligned}
\end{equation}
where $g_t = \sigma (W_g x_t) \in \mathbb{R}$ is a learned scalar gate that determines how much new information overwrites existing information.
They also analogously modify the attention normalizer to incorporate $g_t$.
This rule is problematic as it overwrites all state elements equally without fine-grained control.

\citet{schlag2021linear} proposes improving Eq. \ref{eq:attention-rec-peng} using a Fast Weight Programmer \cite{schmidhuber1992learning} \textbf{delta rule} to forget values associated with the current write key by removing the associated value before adding the new value:
\begin{equation}
\label{eq:attention-rec-delta}
S_{t} = S_{t-1} - g_t S_{t-1} \phi'(k_t)\phi'(k_t)^\intercal + g_t v_t \phi'(k_t)^\intercal
\end{equation}
where $g_t$ is a scalar that defines the extent to which the new value replaces the old value.
To stabilize training, \citet{schlag2021linear} applies \textbf{sum normalization} to feature maps to enforce the outputs to have components that sum to 1 (i.e.,~$\phi'(k_t) = \phi(k_t) / \sum_j^d \phi(k_t)_j$).
This normalization is applied to both key and query, and the output is computed as $y_t = S_t \phi'(q_t) / z_t \phi'(q_t)$.
\citet{schlag2021linear} showed that dropping \textbf{attention normalization} (i.e.,~$y_t = S_t \phi'(q_t)$) works just as well and is redundant when combined with sum normalization.

\subsection{Kernel Feature Map}
One motivation for kernel-based methods is to closely approximate self-attention.
\citet{peng2021random} proposes Random Feature Attention (RFA), which uses random feature maps to produce unbiased estimates of the exponential $\text{sim}(x,y)$ function.
\citet{choromanski2020rethinking} proposes FAVOR+, a random feature map with lower variance.
Instead of rigorously approximating self-attention, several proposals aim simply to maintain positive outputs motivated by the positivity of attention weights.
\citet{katharopoulos2020transformers} proposes the $\phi(x) = \text{ELU}(x) + 1$ \cite{clevert2015fast} feature map.
\citet{schlag2021linear} proposes
Deterministic Parameter-Free Projection (DPFP), 
a feature map that expands $m$ without introducing additional learned parameters.
\citet{kasai2021finetuning} proposes using a simple learned ReLU feature map $\phi(x) = \text{ReLU}(W_\phi x + b_\phi)$ that introduces additional parameters and showed better performance than ELU and RFA.
In this work, we use this ReLU feature map as a baseline for its strong language model fine-tuning results.

\section{Decay Update Rule}
We propose the \textbf{decay update rule}, which replaces the self-attention mechanism in Transformers with decaying fast weights \cite{ba2016using} that evolves with linear dynamics.
We modify the additive update rule (Eq. \ref{eq:attention-rec-add}) by adding a low-rank decay matrix $G_t \in (0, 1)^{d \times m}$ to forget information.
\begin{equation}
\label{eq:srt-rec}
\begin{aligned}
S_{t} = G_t \otimes S_{t-1} + v_t \phi(k_t)^\intercal  \\
G_t = \sigma(W_z x_t + b_z) \sigma(W_f x_t + b_f)^\intercal 
\end{aligned}
\end{equation}
where $W_z \in \mathbb{R}^{d \times d}$, $W_f \in \mathbb{R}^{m \times d}$, $b_z \in \mathbb{R}^d$ and $b_f \in \mathbb{R}^m$ are newly added parameters.
The sigmoid activation $\sigma$ in $G_t$ bounds the output values to ensure\footnote{Without sigmoid, training diverges.} stable linear recurrent dynamics \cite{miller2018stable}.
We initialize $b_z,b_f$ via Uniform Gate Initialization \cite{gu2020improving}, then re-scale pre-trained weight $W_v$ by $1-\sigma(b_z)$. This re-scaling prevents initial iterations from numerical overflows.

Our rule is based on the \textbf{gated rule} \cite{peng2021random} and gated fast weights \cite{schlag2017gated}. Unlike the \textbf{gated rule}, our gate $G_t$ is a learned matrix and not a scalar, which enables more control when decaying $S_t$ as feature size $m$ increases.
Compared to \citet{schlag2017gated,peng2021random}, we do not gate the $v_t \phi(k_t)^{\intercal} $ term as it did not bring performance gains given our initialization scheme.
Unlike the \textbf{delta rule} \cite{schlag2021linear,schlag2020learning}, our rule is a pure element-wise operation, where state dimensions $m$ \textit{do not mix} \cite{laurent2016recurrent,chen2017minimalrnn}.
This can enable more efficient parallelization on GPU \cite{lei2021attention} (Fig. \ref{fig:profiling}).

Finally, we choose a linear projection feature map and remove attention normalization \cite{schlag2020learning} and its associated vector $z_t$ from Eq. \ref{eq:attn-norm}.
\begin{equation}
\phi(x) = W_\phi x,
y_t = S_t \phi(q_t)
\end{equation}
In practice, this feature map can be subsumed into $W_k,W_q$ \cite{kasai2021finetuning} and is equivalent to the identity function (i.e.,~\textbf{no feature map}).

\subsection{Implementation Efficiency}
\label{sec:runtime}
During per-token generation, our proposal has the same $O(dm)$ time and memory complexity as prior kernel-based methods instead of the $O(Td)$ requirement of self-attention.
Fig. \ref{fig:profiling} illustrates the practical benefits of delta and decay rules versus self-attention during generation.
For training, we developed a memory-efficient CUDA kernel\footnote{Our optimized CUDA kernels are available at \url{https://github.com/jenni-ai/T2FW}.} that avoids materializing outer product matrices in memory, similar to \citet{katharopoulos2020transformers}.
However, we must store $S$ for backpropagation due to the addition of $G_t$ in Eq. \ref{eq:srt-rec}, which makes the computation non-reversible \cite{mackay2018reversible}.
Thus, training consumes $O(Tdm)$ memory and $O(T)$ parallel time.
In practice, our implementation consumes less memory than self-attention when $m \ll T$.
This implies that for memory-bound training, delta may be more favorable than our decay rule despite having a slower run-time.
We note this trade off for practitioners to consider and leave memory optimizations for future work.

\section{GPT-2 Fine-Tuning Experiments}
We perform fine-tuning experiments to speed up existing pre-trained Transformers in a similar setting to Transformer-to-RNN (T2R) \cite{kasai2021finetuning}.
We choose GPT-2 small \cite{radford2019language} as our candidate model to fine-tune, as it has a direct scale-up to large models such as GPT-3 \cite{brown2020language}.
To facilitate reproducibility on publicly available datasets, we first fine-tune GPT-2 on The Pile \cite{pile} - a large general corpus of text.
This fine-tuned GPT-2 serves as our baseline \textit{target model} (14.5 PPL, Table \ref{tab:gpt2-scale}). When replacing self-attention, we simply swap the layer with a new recurrence formulation. Following \citet{kasai2021finetuning}, the entire model is fine-tuned.

We explore different update rules by fine-tuning the target \textbf{GPT-2} (Eq. \ref{eq:attention}) model into \textbf{add} (Eq. \ref{eq:attention-rec-add}), \textbf{gated} (Eq. \ref{eq:attention-rec-peng}), \textbf{delta} (Eq. \ref{eq:attention-rec-delta})\footnote{Following \citet{schlag2021linear}, we apply sum normalization to the delta rule. We also initialize $g_t \approx 0.007$ to a small value. Without these changes, the model diverges.} and our proposed \textbf{decay} rule (Eq. \ref{eq:srt-rec}).
In all settings, we hold state capacity $m=4$ constant for fair comparison.
As a \textbf{baseline}, we train all rules with ReLU feature maps \cite{kasai2021finetuning} and attention normalization.
For attention normalization, Eq. \ref{eq:attention-rec-peng} is used for the gated rule and Eq. \ref{eq:attention-rec-add} is used otherwise.
We subsequently ablate attention normalization and ReLU feature maps (to no feature map) to analyze their effects.

\subsection{Results}
\begin{table}
\centering
\begin{tabular}{lllll}
\hline
\textbf{Rule} & \textbf{Baseline} & \textbf{Norm} \xmark & $\boldsymbol \phi$ \xmark \\
\hline
Add   & 22.6 & *    & 20.0 \\
Delta & 21.6 & 20.1 & *    \\
Gated & *    & 19.7 & \textbf{17.1} \\
Decay & N/A    & 18.8 & \textbf{17.1} \\
\hline
\end{tabular}
\caption{
Perplexity (PPL) of different update rule configurations on The Pile validation set.
\textbf{Baseline} uses both attention normalization and ReLU feature maps.
\textbf{Norm} \xmark~means we removed normalization from the baseline.
$\boldsymbol \phi$ \xmark~means we removed both ReLU feature maps and normalization from the baseline.
* indicates divergence.}
\label{tab:gpt2}
\end{table}

\begin{table}
\centering
\begin{tabular}{lll}
\hline
 \textbf{Rule} & $m$ & \textbf{PPL} \\
\hline
 GPT-2 & - & 14.5 \\
 Local Attention & - & 19.4 \\
\hline
Gated & 16 & 16.8 \\
Decay & 16 & 16.1 \\
Gated & 32 & 16.4 \\
Decay & 32 & \textbf{14.6} \\
\hline
\end{tabular}
\caption{Update rules on The Pile validation set as $m$ increases. No normalization or feature maps are used.}
\label{tab:gpt2-scale}
\end{table}

\subsubsection{Normalization Ablation}
We first ablate each update rules' proposed attention normalization method to test its effects (\textbf{Baseline} vs \textbf{Norm} \xmark~in Table \ref{tab:gpt2}).
\citet{peng2021random}'s modification of attention normalization (Eq. \ref{eq:attention-rec-peng}) causes the gated rule to diverge, likely due to division by small $z_t$ values.
Without normalization, the gated rule trained with better stability.
Our results indicate that only the additive rule with ReLU feature maps requires normalization to converge.
Otherwise, normalization is redundant, corroborating \citet{schlag2021linear}'s findings.

\subsubsection{Feature Map Ablation}
We remove attention normalization on all rules, and further investigate if positive feature maps are necessary ($\boldsymbol \phi$ \xmark~in Table \ref{tab:gpt2}).
For each rule, we replace the ReLU feature map \cite{kasai2021finetuning} with a linear projection (i.e.,~no feature map).
This simplification improves performance for all rules\footnote{For the additive rule, we re-scaled the pre-trained value weights $W_v$ by $1/512$ to prevent initial iteration overflow.} except the delta rule, which diverges.
Our results suggest that positive feature maps are only necessary when using attention or sum normalization to avoid small divisors during normalization.
To verify this, we also trained the additive rule with normalization but no feature map and it diverged.

\subsubsection{Update Rule Comparison}
When comparing update rules under their best configuration (Table \ref{tab:gpt2}), the delta rule performs the worse.
Surprisingly, under its best configuration, the simple additive rule outperforms the delta rule.
This could be due to a lack of stable recurrent dynamics \cite{chen2017minimalrnn} during the fine-tuning setting.
Both the gated and decay rules outperform other rules in all configurations under $m=4$.

\subsubsection{Scaling to Larger $m$}
Finally, we explore if we can increase state capacity $m$ to close the performance gap with \textbf{GPT-2} baseline (Table \ref{tab:gpt2-scale}).
We also compare against \textbf{local attention} (window size 32), which is regarded as a strong baseline \cite{xiong2021simple}.
We train larger state capacity variants of the decay and gated rule where $m=\{16,32\}$.
Our results show that the decay rule scales to better performance relative to the gated rule as $m$ increases.
At $m=32$, our decay rule recovers 99\% of GPT-2's performance.

\section{WikiText-103 Fine-Tuning Experiments}
\begin{table}
\centering
\begin{tabular}{llllll}
\hline
\textbf{Model} & \textbf{Memory} & \textbf{Test (0)} & \textbf{Test (480)} \\
\hline
Base & - & 20.5 & 19.0 \\
\hline
†Linformer & $2\times64$ & 27.2 & 30.7 \\
†$\text{ABC}_{\text{MLP}}$& $2\times32$ & 21.9 & 20.5 \\
\hline
Decay & 32 & 22.1 & 20.7 \\
Decay & 64 & 21.9 & 20.5 \\
\hline
\end{tabular}
\caption{
Comparison under \citet{baevski2018adaptive} setting on WikiText-103 test set with context sizes 0 and 480.
† results are from \citet{peng2021abc}, which stores both key and value vectors, doubling the memory size.
We do not compare with memory size $\text{ABC}_{\text{MLP}}$ $2\times64$ as it effectively stores 128 vectors.
}
\label{tab:wik103}
\end{table}

We perform a similar experiment to fine-tune a pre-trained Transformer into our decay rule with $m=\{32,64\}$ on the WikiText-103 \cite{merity2016pointer} language modeling dataset.
We fine-tune from the publicly available checkpoint from \citet{baevski2018adaptive} (\textbf{Base}) by swapping attention with our decay rule and tuning the entire model, with similar hyperparameters as \citet{peng2021abc}.
We compare against recently proposed self-attention approximations including a causal variant of \textbf{Linformer} \cite{wang2020linformer} and Attention with Bounded Memory Control (\textbf{$\text{ABC}_{\text{MLP}}$}) \cite{peng2021abc} evaluated under the same setting of 0 and 480 token context sizes on the WikiText-103 test set.
Our method is simpler than these approaches and performs competitively\footnote{Linformer and ABC are trained from scratch, while our work focuses on the fine-tuning setting.} under similar state memory sizes (Table \ref{tab:wik103}).

\section{Limitations}
To compare update rules, our experiments use the simple feature map from \citet{kasai2021finetuning} for all rules.
However, we acknowledge that some of these rules are jointly proposed with their feature maps, which may be required for good performance.
For example, when training \citet{peng2021random}'s gated rule (\textbf{baseline}), we attempted various initialization schemes and tried adding small constants to the normalization divisor to help the model converge.
However, none of our efforts worked.
We note that \citet{peng2021random}'s original proposal of the gated rule does not use the more recently proposed learned ReLU feature maps \cite{kasai2021finetuning}, which may be incompatible with their original proposal.
Similarly, \citet{schlag2021linear}'s delta rule was also proposed along with their feature map.
Our results suggest that their proposed update rules may not be robust to alternative feature maps or the fine-tuning setting.

The majority of our experiments are performed on The Pile with GPT-2 and assumes a large dataset pre-training setup.
While this setup is typical of many NLG applications, it may not generalize to settings without a large dataset, such as low-resource NLP settings.
In these cases, more complex approaches may provide better inductive biases than our simple approach.

Our work focuses on the specific case of fine-tuning pre-trained auto-regressive Transformer language models into fast recurrent variants for inference applications.
However, we do not explore training these models from scratch or on different architectures.
It may be the case that our approach only works in the fine-tuning setting and that self-attention pre-training is required \cite{kasai2021finetuning}.

Our work also exclusively focuses on language modeling and does not consider evaluation on downstream tasks.
While language modeling performance typically correlates to downstream task performance \cite{brown2020language}, future work should further validate our proposal on these practical areas of interest.

Finally, as mentioned in Sec. \ref{sec:runtime}, our proposed update rule requires more memory (i.e., $O(Tdm)$) than the add and delta rule (i.e., $O(Td)$) during training.
This may lead to memory issues when training state sizes with larger $m$, in which the delta rule may be preferred over the decay rule.
Future work should explore extensions of our approach with better space complexity.

\section{Ethical Considerations}
Our work uses datasets crawled from the public web, including WikiText-103 \cite{merity2016pointer} and The Pile \cite{pile} and may contain sensitive information or contain undesirable biases.
We refer readers to the dataset descriptions from their respective papers for details.

Our work focuses on improving the inference speed of generating from language models by fine-tuning pre-trained models, which may benefit downstream applications that require the deployment of language models.
Our work does not explore the issues of bias and disinformation in language models nor specifically aims to mitigate these issues.
Our models will likely exhibit the same biases and issues that large language models exhibit
\cite{brown2020language}.
Practitioners who deploy models with our proposal should not assume our method mitigates these issues.
Text sampled from these models may contain offensive content and biases and we advise practical uses of these models to involve some form of human supervision.

\bibliography{anthology,custom}

\appendix

\section{GPT-2 Experimental Details}
\label{sec:appendix}
When fine-tuning GPT-2 small on The Pile\footnote{\url{https://pile.eleuther.ai/}}, we started fine-tuning from the publicly available checkpoint provided by Hugging Face\footnote{\url{https://huggingface.co/gpt2}}.
The Pile consists of 825 GB of diverse text (English-dominant) including stories, websites, code and mathematical questions.
It is intended for large-scale language model pre-training.
Due to the scale of The Pile, our fine-tuning only sees a small subset of the training set.
We evaluate on the provided validation set from The Pile.

GPT-2 small \cite{radford2019language} has approximately 124M parameters.
All models are fine-tuned with a batch size of 32 for 100,000 iterations with a learning rate of $6 \times 10^{-4}$ and gradient clipping of 1.0 on a single NVIDIA A100 GPU for 13 hours.
Training was performed on Pytorch 1.11.
We use mixed precision training \cite{micikevicius2017mixed} except for computing normalizer $z_t$ (Eq. \ref{eq:attention-rec-add}), which is prone to numerical overflows.
We evaluate on the validation set every 4000 training iterations and report the best validation perplexity achieved for a single run.

\section{WikiText-103 Experimental Details}
WikiText-103 \cite{merity2016pointer} is an English dataset of articles scraped from the Good and Featured articles on Wikipedia with 103K training, 217K validation and 245K test word tokens.
The dataset is available under the Creative Commons Attribution-ShareAlike License.
Compared to The Pile, WikiText-103 is a significantly smaller dataset with approximately 528 MB of text.
We fine-tune starting from the checkpoint provided by \citet{baevski2018adaptive}\footnote{\url{https://github.com/pytorch/fairseq/tree/main/examples/language_model}}, which has 242M parameters.

Training was performed using the Fairseq library\footnote{\url{https://github.com/pytorch/fairseq}} on Pytorch 1.11.
A few manual hyperparameter searches were performed based on hyperparameters used in \citet{peng2021abc}, which included adjustments to the learning rate and the choice of optimizer.
For both models, we trained with a batch size of 26 for 100,000 iterations.
We used Adam optimizer with 4000 warm-up steps and decayed the learning rate using a cosine schedule to $2 \times 10^{-6}$ for the rest of the training iterations.
We used a maximum learning rate of $10^{-4}$ and gradient clipping of $0.1$.
Training was performed on a single NVIDIA A40 GPU for 35 hours.
We report the test perplexity achieved at the end of the training for a single run.

\end{document}